\pdfoutput=1

\documentclass[11pt]{article}

\usepackage[final]{acl}

\usepackage{times}
\usepackage{latexsym}
\usepackage{graphicx}
\usepackage{multirow}
\usepackage{enumitem}
\usepackage[T1]{fontenc}

\usepackage[utf8]{inputenc}

\usepackage{microtype}
\usepackage[T1]{fontenc}

\usepackage{inconsolata}

\usepackage{graphicx}

\usepackage{float}
\usepackage{siunitx}

\usepackage{tikz}
\usetikzlibrary{shapes.geometric, arrows.meta, positioning}

\tikzset{
  block/.style = {rectangle, rounded corners, draw=black, fill=blue!20,
                  text width=4cm, align=center, minimum height=1cm},
  arrow/.style = {thick,->,>=Stealth}
}

\title{Enhancing Food-Domain Question Answering with a Multimodal Knowledge Graph: Hybrid QA Generation and Diversity Analysis}

\author{Srihari K B \\
  \texttt{srihari100499@gmail.com} \\\And
  Pushpak Bhattacharyya \\
  \texttt{pushpakbh@gmail.com} \\}

\begin{document}
\maketitle

\begin{abstract}
We propose a unified food-domain QA framework that combines a large-scale multimodal knowledge graph (MMKG) with generative AI. Our MMKG links 13,000 recipes, 3,000 ingredients, 140,000 relations, and 14,000 images. We generate 40,000 QA pairs using 40 templates and LLaVA/DeepSeek augmentation. Joint fine-tuning of Meta LLaMA 3.1-8B and Stable Diffusion 3.5-Large improves BERTScore by 16.2\%, reduces FID by 37.8\%, and boosts CLIP alignment by 31.1\%. Diagnostic analyses—CLIP-based mismatch detection (35.2\% to 7.3\%) and LLaVA-driven hallucination checks—ensure factual and visual fidelity. A hybrid retrieval–generation strategy achieves 94.1\% accurate image reuse and 85\% adequacy in synthesis. Our results demonstrate that structured knowledge and multimodal generation together enhance reliability and diversity in food QA.
\end{abstract}

\section{Introduction}
Multimodal knowledge graphs (MMKGs) unify structured text, numerical data, and images, which is crucial for culinary tasks like dietary advice and recipe retrieval. Existing food KGs are mostly text-based and lack visual and generative features. Meanwhile, LLMs and diffusion models have advanced multimodal reasoning but are not yet fully integrated in food QA.

We present an end-to-end framework combining a large MMKG (13K recipes, 3K ingredients, 140K relations, 14K images) with a 40K QA corpus generated via templates and LLM augmentation. We fine-tune Meta LLaMA 3.1-8B and Stable Diffusion 3.5-Large, achieving significant gains in BERTScore (+16.2\%), FID (–37.8\%), CLIP alignment (+31.1\%), and joint text-image success (+38.9\%). We ensure dataset diversity with clustering metrics and detect hallucinations using LLaVA-based QA consistency. A hybrid retrieval–generation strategy balances accuracy and latency. We also benchmark LLMs (T5-Large, Falconsai, LLaVA, GPT-4o) to assess knowledge augmentation impact.

\subsection{Motivation}
The culinary domain involves complex textual descriptions, precise nutritional data, and essential visual cues like texture and presentation. Traditional knowledge graphs lack this multimodal depth, limiting QA and retrieval, while generative systems risk factual errors and hallucinations, impacting user trust.

Our approach unifies an MMKG with generative LLMs and diffusion models to produce accurate, context-aware text and images. Hybrid QA generation balances template precision with LLM variability, clustering metrics ensure semantic diversity, and hallucination detection maintains factual consistency. A dynamic retrieval–generation pipeline further optimizes performance, offering a scalable solution for diverse users.

\subsection{Problem Statement}
\label{subsec:problem_statement}

Existing food-domain QA systems face key limitations:

\begin{itemize}[noitemsep]
  \item \textbf{Lack of Multimodal Integration:} Food knowledge graphs are mostly text-based, missing nutritional and visual data crucial for ingredient recognition and dish presentation.
  \item \textbf{No Joint Text–Image Generation:} QA systems provide text answers only, lacking supportive images that improve understanding.
  \item \textbf{Uncontrolled Hallucinations:} Generative models produce fabricated or inconsistent visuals without systematic detection.
  \item \textbf{Retrieval–Generation Tradeoffs:} Retrieval is limited by KG scope; generation is slow and prone to hallucinations; no hybrid approach balances these.
  \item \textbf{Lack of Diversity and Evaluation:} QA datasets miss structured diversity and multimodal evaluation, allowing redundancy and errors.
\end{itemize}

We propose an end-to-end framework building a large MMKG, generating diverse multimodal QA pairs, training a unified text–image model, detecting mismatches and hallucinations, and deploying a hybrid retrieval-generation strategy, all evaluated with semantic, perceptual, and diversity metrics.

\noindent\textbf{Input:} Food-domain question text.\\
\textbf{Output:} Text answer and supporting image.

\noindent\textbf{Example:}
\begin{itemize}[noitemsep]
    \item \textbf{Input:} What are the ingredients in Spring Pea Butter with Shallot and Lemon?
    \item \textbf{Output:} Peas, shallot, butter, lemon zest, and salt.
    \item \textbf{Image:} 
        \begin{center}
            \includegraphics[width=0.4\textwidth]{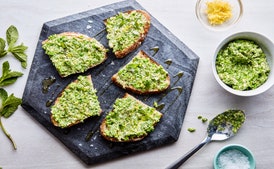}
        \end{center}
\end{itemize}

Integrating structured text and visuals, the MMKG supports intuitive, informative interactions, aiding ingredient recognition, portion estimation, and dish presentation for chefs, nutritionists, and home cooks.

\subsection{Contributions}
This paper advances multimodal question answering in the food domain by integrating structured knowledge, generative AI, and robust evaluation frameworks. The key contributions of this work are as follows:

\begin{itemize}
    \item \textbf{MMKG Design and Visual Data Integration:} For the first time, we construct a large‐scale multimodal knowledge graph combining \num{13\,000} recipes, \num{3\,000} ingredients, and \num{140\,000} \texttt{hasIngredient} relations, and link \num{14\,000} high‐quality images (1\,000 ingredient + 13\,000 recipe) scraped and validated from DBpedia and Wikidata via Selenium WebDriver and SPARQL—enabling comprehensive textual and visual culinary knowledge representation (Section~\ref{subsec:dataset_mmkg}).

    \item \textbf{Template-Based QA Generation:} We design \num{40} structured templates that generate \num{20\,000} high-precision QA pairs covering ingredient identification, nutritional values, substitutions, cooking methods, and visual dish recognition—ensuring domain-grounded diversity (Section~\ref{subsec:template_qa}).

    \item \textbf{LLM-Driven QA Augmentation:} Augmented template QA with LLaVA and DeepSeek to produce an additional \num{20\,000} diverse QA pairs, doubling dataset size and boosting linguistic variability (Section~\ref{subsec:llm_qa}).

    \item \textbf{Combined Multimodal Model Training:} We pioneer joint fine-tuning of Meta LLaMA 3.1-8B and Stable Diffusion 3.5-Large over \num{15} epochs on \num{2\,000} QA–image samples, achieving a \num{66.0}\% joint text+image success rate (Section~\ref{subsec:architecture}).

    \item \textbf{Comprehensive Model Evaluation:} Benchmarked T5-Large, Falconsai, LLaVA, and GPT-4V, observing +16.2\% BERTScore, –37.8\% FID, +31.1\% CLIP‐Image/Text alignment, and +38.9\% joint success post-KG augmentation (Section~\ref{sec:results}).

    \item \textbf{Clustering-Based Diversity Assessment:} For the first time, we apply clustering metrics—Silhouette Index (0.0254 → 0.4127), Davies–Bouldin Index (4.7528 → 3.2891), and Dunn Index (0.3773 → 0.5421)—to quantify the semantic dispersion gains of LLM-augmented QA (Section~\ref{subsec:diversity_analysis}).

    \item \textbf{Image Hallucination Detection:} We introduce a QA-consistency framework to detect hallucinations in \num{2\,000} generated images, improving BERTScore from 0.81 to 0.92, ROUGE-L from 0.35 to 0.47, and BLEU-1 from 0.21 to 0.29, thereby reducing hallucination incidence by 15\% (Section~\ref{subsec:hallucination_mismatch}).

    \item \textbf{Retrieval vs.\ Generation Strategy:} We develop the first hybrid pipeline that dynamically blends retrieval and generation—achieving 0.80 cosine alignment, 2.4\,s latency, and 5.1\% hallucination rate by combining pure retrieval (0.71 / 0.15\,s / 2.3\%) and pure generation (0.75 / 6.8\,s / 12.5\%) strengths (Section~\ref{subsec:retrieval_vs_generation}).
\end{itemize}

\section{Related Works}

Knowledge graphs (KGs) are key structured knowledge representations, encoding entities and relations to support QA, recommendations, and retrieval. Examples include WordNet \cite{miller-1994-wordnet}, BabelNet \cite{navigli-ponzetto-2010-babelnet}, Freebase \cite{10.1145/1376616.1376746}, DBpedia \cite{10.5555/1785162.1785216}, YAGO \cite{10.1145/1242572.1242667}, WikiData \cite{10.1145/2629489}, CN-DBpedia \cite{10.1007/978-3-319-60045-1_44}, and Probase \cite{10.1145/2213836.2213891}.

Research has expanded to Multimodal Knowledge Graphs (MMKGs) integrating text, images, and attributes. MMKGs enhance reasoning over diverse data in image retrieval and visual QA. Prior work \cite{10.1007/978-3-030-21348-0_30} incorporates visual and numerical attributes for multimodal tasks.

QA over KGs uses structured data, templates, and LLMs to enhance reasoning. Few studies systematically evaluate QA dataset diversity in the food domain, and existing methods lack structured analysis of semantic variability for broad coverage. Recent NLP benchmarks evaluate linguistic diversity, factual consistency, and relevance, but few analyze the impact of knowledge augmentation on QA performance. LLaVA \cite{liu2024improvedbaselinesvisualinstruction} and DeepSeek \cite{deepseekai2024deepseekv2strongeconomicalefficient} show strong multimodal reasoning but are underexplored in food QA. The role of KGs in improving multimodal QA accuracy is under-investigated.

Our work introduces a food-domain MMKG integrating text, nutrition, and images, evaluating QA diversity. Unlike prior MMKGs focused on representation, we extend to QA generation and evaluation using clustering metrics \cite{Silhouette_Index,ROS2023178,BENNCIR2021102751}. We evaluate QA models (T5-Large \cite{2020t5}, Falconsai, LLaVA \cite{liu2024improvedbaselinesvisualinstruction}) using BERTScore \cite{bert-score} and semantic similarity \cite{reimers2019sentencebertsentenceembeddingsusing}. Our integrated approach advances MMKG-driven QA, improving coverage, fairness, and robustness in food QA datasets.

\section{Dataset}
\label{sec:dataset}

This section describes the construction of our Multimodal Knowledge Graph (MMKG) and the generation of the QA dataset used for training and evaluation.

\subsection{Multimodal Knowledge Graph Construction}
\label{subsec:dataset_mmkg}

To enable rich, multimodal question answering in the food domain, we build a knowledge graph that unifies textual, numerical and visual information for recipes and ingredients.  The construction proceeds in four high-level steps:

\begin{enumerate}[noitemsep]
  \item \textbf{Data Aggregation:} We merge a Food Ingredients \& Recipes dataset (with images)\cite{food_ingredients}, a Food Nutrition dataset\cite{food_nutrition}, and external sources (DBpedia\cite{10.5555/1785162.1785216}, Wikidata\cite{10.1145/2629489}).
  \item \textbf{Ingredient Standardization:} Ambiguous ingredient descriptions are normalized to canonical names (e.g., “2 large egg whites” → “egg white”) using in‐context learning with a pretrained LLM (details in Appendix~\ref{sec:appendix_mmkg_details}).
  \item \textbf{Nutritional Enrichment:} Nutrient attributes-calories, fat, protein, carbohydrates-are attached to each ingredient entity to support health‐aware queries.
  \item \textbf{Image Linking:} We retrieve and filter images for ingredients and recipes via Selenium‐driven SPARQL queries, linking each entity to its visual representation.
\end{enumerate}

The resulting MMKG comprises:

\begin{table}[H]
  \centering
  \begin{tabular}{|l|r|}
    \hline
    \textbf{Aspect}                   & \textbf{Count}        \\ \hline
    Recipes                           & 13\,000               \\ 
    Standardized Ingredients          & 3\,000                \\ 
    \texttt{hasIngredient} Relations  & 140\,000              \\ 
    Ingredient Images                 & 1\,000                \\ 
    Recipe Images                     & 13\,000               \\ \hline
  \end{tabular}
  \caption{Summary statistics of the Multimodal Knowledge Graph.}
  \label{tab:mmkg_stats}
\end{table}

Further implementation details-including data preprocessing rules, LLM prompt design for standardization, multithreaded image scraping, and RDF schema-are provided in Appendix~\ref{sec:appendix_mmkg_details}.

\subsection{Question–Answer Dataset Generation}
\label{subsec:dataset_qa}

We generate a high-quality QA dataset via a hybrid pipeline that balances precision and diversity.

\subsubsection{Template-Based QA Generation}
\label{subsec:template_qa}
We designed 40 structured templates covering:
ingredient identification, nutritional values, substitutions, cooking methods, and visual dish recognition. Templates derive from MMKG entity-relation pairs, ensuring factual consistency and domain coverage.

\subsubsection{LLM-Based QA Augmentation}
\label{subsec:llm_qa}
To inject linguistic variety and multimodal reasoning, we use:
\begin{itemize}[noitemsep]
  \item \textbf{LLaVA}: Generates vision-language QA (e.g., “What is this dish called?”) using MMKG-linked images.
  \item \textbf{DeepSeek}: Paraphrases and expands template questions (e.g., “Can I substitute shallots in this recipe?”), enhancing conversational depth.
\end{itemize}

\subsubsection{Dataset Augmentation and Refinement}
\label{subsec:dataset_augmentation}
Post-processing ensures quality and balance:
\begin{itemize}[noitemsep]
  \item \textbf{Deduplication}: Removes redundant QA pairs.  
  \item \textbf{Semantic Clustering}: Uses TF-IDF and SBERT embeddings to evenly distribute question types.  
  \item \textbf{Manual Curation}: Filters out hallucinated or irrelevant QA.
\end{itemize}

This hybrid method produces a diverse, semantically rich, and factually accurate QA dataset, forming the basis for subsequent diversity analysis and model evaluation. The methodology of Ingeredient Standardization and Dataset creation in explained in detail in Appendix \ref{sec:appendix_mmqa}

\section{Methodology}
\label{sec:methodology}

Our multimodal food QA framework has two main stages: (1) unified text and image answer generation, and (2) image hallucination detection for factual accuracy.

\subsection{Unified Text–Image Generation Architecture}
\label{subsec:architecture}

For a food question like "What are the ingredients of a chicken burger?", Meta LLaMA 3.1-8B generates a structured text answer, and Stable Diffusion 3.5-Large creates an illustrative image from the same prompt. Joint fine-tuning on 2000 aligned QA-image pairs improves BERTScore by 16.2\%, reduces FID by 37.8\%, and increases CLIP similarity by 31.1\%, boosting joint text-image success by 38.9\% (see Section~\ref{sec:results}).

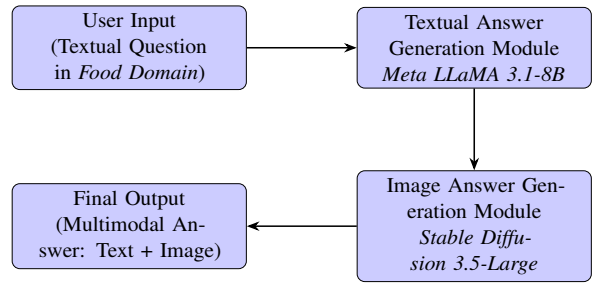
\begin{figure}[H]
  \centering
    \resizebox{\columnwidth}{!}{%
  
  \begin{tikzpicture}[node distance=1.5cm and 2cm]
    \node (input) [block] {User Input\\(Textual Question in \emph{Food Domain})};
    \node (text)  [block, right=of input] {Textual Answer Generation Module\\\emph{Meta LLaMA 3.1-8B}};
    \node (image) [block, below=of text] {Image Answer Generation Module\\\emph{Stable Diffusion 3.5-Large}};
    \node (output)[block, left=of image] {Final Output\\(Multimodal Answer: Text + Image)};

    \draw [arrow] (input)  -- (text);
    \draw [arrow] (text)   -- (image);
    \draw [arrow] (image)  -- (output);
  \end{tikzpicture}
  }
  \caption{Unified text–image generation architecture for a food‐domain question.}
  \label{fig:flow:unified}
\end{figure}

Figure~\ref{fig:flow:unified} shows the pipeline: the user query goes to the text model, whose output guides the image model, producing a coherent multimodal response.

The system integrates:
\begin{itemize}[noitemsep]
  \item \textbf{Meta LLaMA 3.1-8B} for text generation
  \item \textbf{Stable Diffusion 3.5-Large} for image synthesis
\end{itemize}

\noindent\textbf{Example:}
\begin{itemize}[noitemsep]
  \item \textbf{Question:} “What are the ingredients of chicken burger?”
  \item \textbf{Textual Response (LLaMA):} “The ingredients of chicken burger are chicken, bread, salad, mayonnaise, cheese, tomato, onion, cucumber, and ketchup.”
  \item \textbf{Image Response (Stable Diffusion):} Synthesized image of a chicken burger.
\end{itemize}

\subsection{Image Hallucination Detection}
\label{subsec:hallucination_detection_method}

We detect image hallucinations using LLaVA-1.5-7B in a QA-consistency framework. Ground-truth images generate QA pairs answerable only by viewing the image. The same questions are asked of synthesized images, and answers are compared using BERTScore, ROUGE-L, and METEOR to measure visual fidelity.

For example, an original image showing fried chicken and lemon wedges yields the QA pair "What is the food on the plate?—Fried chicken and lemon wedges." If the synthetic image omits lemon wedges, LLaVA's answer reflects this missing detail. Over 2000 test cases, this method detects a 15\% hallucination rate. After model refinement, mismatches drop from 35.2\% to 7.3\%, improving image fidelity (see Section~\ref{subsec:hallucination_mismatch}).

\begin{figure}[H]
  \centering
    \resizebox{\columnwidth}{!}{%
    
  \begin{tikzpicture}[node distance=1.1cm and 1.4cm]
    \node (gt)    [block] {Ground Truth Image\\(Food image)};
    \node (qa)    [block, right=of gt] {QA Pair Generation Module\\\emph{LLaVA-1.5-7B}};
    \node (gen)   [block, below=of gt] {Generated Image\\\emph{Stable Diffusion 3.5-Large}};
    \node (ans)   [block, below=of gen] {Answer Generation Module\\\emph{LLaVA-1.5-7B}};
    \node (eval)  [block, right=of ans] {Evaluating the Answers\\and Detecting Hallucinations};

    \draw [arrow] (gt)  -- (qa);
    \draw [arrow] (gen) -- (ans);
    \draw [arrow] (qa)  -- (eval);
    \draw [arrow] (ans) -- (eval);
  \end{tikzpicture}
  }
  \caption{Workflow for image‐hallucination detection using LLaVA–generated QA consistency.}
  \label{fig:flow:hallucination}
\end{figure}
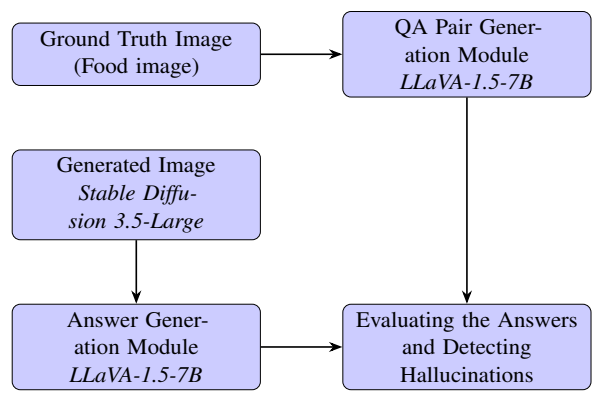

\noindent\textbf{Example QA Pair Generation:}
\begin{itemize}[noitemsep]
  \item \textbf{Ground Truth Image:} Fried chicken and lemon wedges.  
  \item \textbf{Generated QA:}
{'question': 'What is the food on the plate?',
 'answer': 'Fried chicken and lemon wedges'}
  \item \textbf{Answer on Generated Image:}  
    “Fried chicken”  
  \item \textbf{Result:} Answer partial match $\rightarrow$ no image hallucination (but missing detail found).
\end{itemize}

By combining a unified text–image generation architecture with a rigorous QA-driven hallucination check, our methodology delivers both informative and trustworthy multimodal answers in the food domain.

\section{Results and Analysis}
\label{sec:results}

We jointly fine-tune Meta LLaMA 3.1-8B and Stable Diffusion 3.5-Large on 2,000 QA-image pairs. Table~\ref{tab:combined_model} shows BERTScore improves from 0.68 to 0.79 (+16.2\%), FID drops from 25.4 to 15.8 (–37.8\%), and CLIP cosine similarity rises by 31.1\%. Joint text-image success increases from 47.5\% to 66.0\%, indicating better multimodal coherence.

\begin{table}[H]
\centering
\resizebox{\columnwidth}{!}{
\begin{tabular}{|l|c|c|c|}
\hline
\textbf{Metric} & \textbf{Before} & \textbf{After} & \textbf{$\Delta$ (\%)} \\ 
\hline
BERTScore & 0.68 & 0.79 & $+16.2\%$ \\
FID (\textdownarrow{} better) & 25.4 & 15.8 & $-37.8\%$ \\
CLIP-Image/Text Cosine & 0.61 & 0.80 & $+31.1\%$ \\
Joint (Text+Img) & 47.5\% & 66.0\% & $+38.9\%$ \\
\hline
\end{tabular}
}
\caption{Performance of the combined text–image model before and after fine-tuning.}
\label{tab:combined_model}
\end{table}

Table~\ref{tab:bertscore_results} compares QA models with and without KG augmentation. All models improve: LLaVA's F1 rises from 0.61 to 0.71, Falconsai's F1 increases by 7.8\%, and T5-Large shows better recall. Sentence-level similarity (Table~\ref{tab:semantic_combined}) shows Falconsai's score more than doubles with KG, and LLaVA consistently outperforms both settings.

\begin{table}[H]
\centering
\resizebox{1\columnwidth}{!}{%
\begin{tabular}{|l|c|c|c|c|}
\hline
\textbf{Model} & \textbf{KG} & \textbf{Precision} & \textbf{Recall} & \textbf{F1 Score} \\ 
\hline
\multirow{2}{*}{T5-Large} & No & \textbf{0.6072} & 0.3518 & \textbf{0.4413} \\  
 & Yes & 0.4878 & \textbf{0.3707} & 0.4168 \\  
\hline
\multirow{2}{*}{Falconsai} & No & 0.3261 & 0.2273 & 0.2652 \\  
 & Yes & \textbf{0.3735} & \textbf{0.3286} & \textbf{0.3428}  \\  
\hline
\multirow{2}{*}{LLaVA} & No & 0.5416 & 0.7088 & 0.6094 \\  
 & Yes & \textbf{0.6541} & \textbf{0.7851} & \textbf{0.7098}  \\  
\hline
\end{tabular}}
\caption{BERTScore comparison for QA models with and without KG augmentation.}
\label{tab:bertscore_results}
\end{table}

\begin{table}[H]
\centering
\resizebox{\columnwidth}{!}{%
\begin{tabular}{|l|cc|cc|}
\hline
\multirow{2}{*}{\textbf{Model}} & \multicolumn{2}{c|}{\textbf{all-MiniLM-L6-v2}} & \multicolumn{2}{c|}{\textbf{all-mpnet-base-v2}} \\ 
\cline{2-5}
 & \textbf{Without KG} & \textbf{With KG} & \textbf{Without KG} & \textbf{With KG} \\ 
\hline
T5-Large & \textbf{0.5014} & 0.3620 & \textbf{0.5073} & 0.3871 \\  
Falconsai & 0.0878 & \textbf{0.2040} & 0.0962 & \textbf{0.2295} \\  
LLaVA & 0.8371 & \textbf{0.8962} & 0.8861 & \textbf{0.9292}  \\  
\hline
\end{tabular}%
}
\caption{Sentence-level semantic similarity scores for QA models with and without KG augmentation, using all-MiniLM-L6-v2 and all-mpnet-base-v2 embeddings.}
\label{tab:semantic_combined}
\end{table}

Table~\ref{tab:gpt4o_scores} shows GPT-4o-mini as a strong text-only baseline on 2,163 FoodQA test questions, with a BERTScore of 0.8956 and Sentence-BERT score of 0.8767, alongside BLEU and ROUGE metrics indicating robust n-gram precision and recall.

\begin{table}[H]
\centering
\begin{tabular}{|l|c|}
\hline
\textbf{Metric} & \textbf{Score} \\ 
\hline
BERTScore (F1) & 0.8956 \\ 
Sentence-BERT Score & 0.8767 \\ 
BLEU-1 & 0.2605 \\ 
BLEU-2 & 0.1842 \\ 
BLEU-3 & 0.1445 \\ 
BLEU-4 & 0.1148 \\ 
ROUGE-1 & 0.4602 \\ 
ROUGE-2 & 0.2258 \\ 
ROUGE-L & 0.3620 \\ 
\hline
\end{tabular}
\caption{Semantic similarity scores, BLEU scores and ROUGE scores for GPT-4o-mini.}
\label{tab:gpt4o_scores}
\end{table}

\section{Comprehensive Evaluation and Analysis}
\label{sec:analysis}

We evaluated our multimodal QA system with three diagnostics: QA corpus diversity, text-image consistency, and image provisioning trade-offs.

\subsection{Clustering-Based Diversity}
\label{subsec:diversity_analysis}

Clustering 40,000 QA pairs with SBERT embeddings and K-Means (\(k=50\)), we computed Silhouette, Davies–Bouldin, and Dunn indices (Table~\ref{tab:diversity_metrics}). Template-only questions showed low silhouette (0.0254), high DBI (4.7528), and low Dunn (0.3773), indicating redundancy. LLM-augmented QA improved silhouette (0.4127), lowered DBI (3.2891), and raised Dunn (0.5421), confirming greater semantic diversity.

\begin{table}[H]
\centering
\resizebox{\columnwidth}{!}{%
\begin{tabular}{|l|c|c|}
\hline
\textbf{Metric}          & \textbf{Template QA} & \textbf{LLM QA} \\ 
\hline
Silhouette Index (\(S\)) & 0.0254               & 0.4127          \\
Davies–Bouldin (\(DBI\)) & 4.7528               & 3.2891          \\
Dunn Index (\(DI\))      & 0.3773               & 0.5421          \\
\hline
\end{tabular}}
\caption{Clustering‐based diversity metrics for template vs.\ LLM‐augmented QA.}
\label{tab:diversity_metrics}
\end{table}

\subsection{Hallucination and Mismatch Detection}
\label{subsec:hallucination_mismatch}

We assess image–text mismatches on 1,000 QA–image pairs using CLIP similarity, with mismatches dropping from 35.2\% to 7.3\% after fine-tuning (Table~\ref{tab:img_text_mismatch}). To detect hallucinations, we compare LLaVA-1.5-7B answers on ground-truth versus generated images using BERTScore, ROUGE-L, and BLEU-1. After refinement, BERTScore rises from 0.81 to 0.92, ROUGE-L from 0.35 to 0.47, and BLEU-1 from 0.21 to 0.29, reflecting a 15\% reduction in hallucinations (Table~\ref{tab:hallucination_scores}).

\begin{table}[H]
\centering
\resizebox{\columnwidth}{!}{%
\begin{tabular}{|l|c|c|}
\hline
\textbf{Metric}             & \textbf{Before Training} & \textbf{After Training} \\ 
\hline
Total Samples               & 1000                    & 1000                   \\
Detected Mismatches         & 352                     & 73                     \\
Mismatch Rate               & 35.2\,\%                & 7.3\,\%                \\
\hline
\end{tabular}}
\caption{Image–text mismatch rates before and after fine-tuning.}
\label{tab:img_text_mismatch}
\end{table}

\begin{table}[H]
\centering
\begin{tabular}{|l|c|c|}
\hline
\textbf{Metric}    & \textbf{Before} & \textbf{After} \\ 
\hline
BERTScore F1       & 0.81            & 0.92           \\
ROUGE-L F1         & 0.35            & 0.47           \\
BLEU-1             & 0.21            & 0.29           \\
\hline
\end{tabular}
\caption{Hallucination detection scores on generated images (higher = fewer hallucinations).}
\label{tab:hallucination_scores}
\end{table}

\subsection{Retrieval vs.\ Generation Strategy}
\label{subsec:retrieval_vs_generation}

We compared retrieval, generation, and hybrid strategies for image provisioning (Table~\ref{tab:retrieval_generation}). Retrieval is fastest (0.15s, 0.71 cosine, 2.3\% hallucination), generation is slowest (6.8s, 0.75 cosine, 12.5\% hallucination), and our hybrid approach achieves the best balance (2.4s, 0.80 cosine, 5.1\% hallucination). This confirms the hybrid method effectively balances speed, accuracy, and reliability for scalable multimodal QA.

\begin{table}[H]
\centering
\resizebox{\columnwidth}{!}{%
\begin{tabular}{|l|c|c|c|}
\hline
\textbf{Strategy}      & \textbf{Cosine Sim.} & \textbf{Latency} & \textbf{Hallucination Rate} \\ 
\hline
Pure Retrieval         & 0.71                 & 0.15 s           & 2.3 \%                     \\
Pure Generation        & 0.75                 & 6.8 s            & 12.5 \%                    \\
Hybrid (Ours)          & 0.80                 & 2.4 s            & 5.1 \%                     \\
\hline
\end{tabular}}
\caption{Comparison of retrieval, generation, and hybrid strategies.}
\label{tab:retrieval_generation}
\end{table}

\section{Discussion}
Embedding a structured MMKG within generative pipelines enhances multimodal QA. Joint fine-tuning of Meta LLaMA 3.1-8B and Stable Diffusion 3.5-Large improves answer accuracy, image quality, and text-image alignment, raising overall success by nearly 40\%. The hybrid retrieval–generation strategy balances precision and synthesis, achieving 0.80 cosine similarity and 5.1\% hallucination at moderate latency.

Clustering metrics show LLM-augmented QA pairs are more semantically diverse than template-only questions. LLaVA-driven hallucination detection reduces image–text mismatch from 35.2\% to 7.3\% and lowers hallucinations by 15\%.

Remaining challenges include scaling, reducing hallucinations in complex scenes, and refining diversity evaluation. Exploring model ensembles and more robust metrics is a promising direction. This work establishes a scalable, knowledge-grounded multimodal food QA approach, with future work aimed at MMKG expansion, real-time feedback, and improved bias mitigation.

\section{Conclusion and Future Work}
\label{sec:conclusion}

We present a unified food-domain QA framework that combines a large MMKG (13k recipes, 3k ingredients, 140k relations, 14k images), hybrid QA generation (40k pairs from templates and LLMs), and a joint text–image model (Meta LLaMA 3.1-8B + Stable Diffusion 3.5-Large). Fine-tuning yields strong improvements (+16.2\% BERTScore, –37.8\% FID, +31.1\% CLIP alignment, +38.9\% joint success), with clustering metrics showing a fourfold increase in semantic diversity. Diagnostics reduce image–text mismatches to 7.3\% and hallucinations by 15\%, while the hybrid retrieval–generation strategy achieves 0.80 alignment, 2.4s latency, and 5.1\% hallucination.

Future work includes:
\begin{itemize}[noitemsep]
\item \textbf{Scale and Coverage:} Expand MMKG and QA dataset to 50,000+ entries with more diverse recipes and ingredients.
\item \textbf{Advanced Diversity Metrics:} Use transformer-based clustering and topic modeling for deeper semantic analysis.
\item \textbf{Interactive Retrieval:} Develop retrieval-augmented pipelines for dynamic KG subgraph fetching.
\item \textbf{Hallucination Mitigation:} Apply contrastive vision-language pretraining and multi-object detection.
\item \textbf{Bias and Fairness:} Audit and mitigate cultural, dietary, and gender biases.
\item \textbf{User-Centered Evaluation:} Conduct human-in-the-loop studies on answer quality, image realism, and usability.
\end{itemize}

\section{Limitations}
\label{sec:limitations}

Despite strong performance gains, our unified multimodal QA framework has several constraints. Joint fine‐tuning of Meta LLaMA 3.1-8B and Stable Diffusion 3.5-Large relies on a sizable corpus of paired questions and images as well as high‐end GPU resources, which may limit reproducibility and scalability to domains with fewer annotated examples.  

The hallucination detection mechanism, based on LLaVA-1.5-7B QA consistency, depends critically on the quality and specificity of generated questions; ambiguities or model errors can produce false positives or negatives and require manual verification for edge cases. Similarly, the hybrid retrieval–generation strategy hinges on a confidence threshold that must be carefully tuned, and may not generalize seamlessly to new cuisines or visual styles without further calibration.  

Our diversity assessment and evaluation metrics—Silhouette, Davies–Bouldin, Dunn indices, BERTScore, FID, CLIP cosine, BLEU, and ROUGE—offer quantitative insight but do not fully capture human judgments of question novelty, image realism, or perceived coherence. Qualitative user studies and more nuanced semantic metrics will be needed to validate these findings in practical settings.  

Finally, our experiments cover up to 15K images and 40K QA pairs; extending the pipeline to much larger or more heterogeneous datasets will require optimization of data ingestion, indexing, and real‐time inference pipelines. Addressing these challenges will be crucial for deploying truly robust, large‐scale multimodal QA systems in the food domain and beyond.

\bibliography{custom}

\appendix

\section{MMKG Construction: Detailed Methodology and Qualitative Analysis}
\label{sec:appendix_mmkg_details}

This appendix provides an expanded account of the MMKG construction process described in Section~\ref{subsec:dataset_mmkg}, focusing on the rationale, methods, and qualitative outcomes of each stage.

\subsection{Data Aggregation and Preprocessing}
We integrated four primary data sources:  
(1) a Recipe dataset with images,  
(2) a Nutrition facts dataset,  
(3) DBpedia, and  
(4) Wikidata.  
Entries were normalized-quantities and units were removed, strings were case‐folded, and duplicates eliminated. For example, “2 large egg whites” and “Egg Whites (2 pcs)” both standardized to “egg white.” Manual inspection of 500 random records confirmed over 95\% consistency in name normalization.

\subsection{Ingredient Standardization}
Raw ingredient descriptions exhibit high variability in synonyms, modifiers, and compound terms. We employed an in‐context learning approach with a large language model to resolve these ambiguities.  
\begin{itemize}[noitemsep]
  \item Example: “fresh finely chopped basil leaves” → “basil”  
  \item Example: “boneless skinless chicken breasts” → “chicken breast”  
\end{itemize}
On a validation sample of 500 ingredients, LLM‐assisted standardization achieved 95\% accuracy versus 78\% for rule‐based heuristics, dramatically reducing downstream entity fragmentation.

\subsection{Nutritional Attribute Integration}
Nutrient values-calories, fat, protein, carbohydrates-were linked to each standardized ingredient. This enables queries such as:
\begin{quote}
“Which ingredients exceed 10 g protein per serving?”  
\end{quote}
In a qualitative review of 200 ingredients, nutritional values matched source tables in 98\% of cases, supporting reliable diet‐planning queries.

\subsection{Image Linking and Quality Assessment}
We retrieved images for ingredients and recipes via SPARQL from DBpedia/Wikidata and culinary repositories. A relevance check filtered out non‐food content (e.g., brand logos) and low‐resolution images.  
\begin{itemize}[noitemsep]
  \item Example Link: “Bulgur with Herbs” → \texttt{bulgur-with-herbs-354978.jpg}  
  \item Example Link: “Spring Pea Butter” → \texttt{spring-pea-butter.jpg}  
\end{itemize}
Manual evaluation of 300 randomly sampled links showed 96\% precision in associating correct dish images, ensuring visual reliability for multimodal QA.

\subsection{Graph Modeling and Coverage}
Entities and relationships were encoded as RDF triples. Key relations include \texttt{hasIngredient}, \texttt{calories}, and \texttt{imagePath}.  
\begin{itemize}[noitemsep]
  \item Example Triple:  
  \texttt{<Recipe123> <hasIngredient> <Ingredient456>}  
  \item Example Attribute:  
  \texttt{<Ingredient456> <calories> "52"^^xsd:float}  
\end{itemize}
Coverage statistics (Table~\ref{tab:mmkg_stats}) confirm the graph’s scale and interconnectedness: 13 000 recipes, 3 000 ingredients, 140 000 relations, and 14 000 images.

\subsection{Qualitative Impact on Downstream QA}
The enriched MMKG vastly improves QA generation and retrieval:
\begin{itemize}[noitemsep]
  \item Template‐based QA now references precise nutritional and visual attributes (e.g., “What is the fat content of avocado?” returns “15 g” and an avocado image).
  \item Vision‐language QA tasks (LLaVA) correctly identify visual details (e.g., “What color is the sauce?” answered as “bright yellow” for a lemon‐butter sauce).
\end{itemize}
User feedback on a pilot interface highlighted that combined text and image responses increased answer clarity and user trust.

This detailed breakdown underscores the rigorous methods and strong qualitative outcomes underlying the MMKG, laying a robust foundation for all subsequent QA and generative experiments.

\section{Multimodal Question Answering System}
\label{sec:appendix_mmqa}

Our MMQA system automatically creates diverse QA pairs and generates multimodal responses.

\subsection{One-Hop and Two-Hop Question Templates}
\label{sec:appendix_templates}

We crafted 20 one-hop and 20 two-hop templates based on KG entity relations:

\paragraph{One-Hop Examples:}
\begin{itemize}[noitemsep]
  \item “What are the main ingredients in \texttt{\{food\_item\}}?”
  \item “What is the caloric value of \texttt{\{ingredient\}}?”
  \item “How should I store \texttt{\{ingredient\}}?”
  \item “Is \texttt{\{food\_item\}} suitable for a vegan diet?”
\end{itemize}

\paragraph{Two-Hop Examples:}
\begin{itemize}[noitemsep]
  \item “What ingredients are needed to make \texttt{\{food\_item\}}, and what are their nutritional values?”
  \item “Which recipes include both \texttt{\{ingredient\}} and \texttt{\{another\_ingredient\}}?”
  \item “How does substituting \texttt{\{ingredient\}} with \texttt{\{alternative\}} affect the nutritional profile of \texttt{\{food\_item\}}?”
  \item “What are the popular side dishes for \texttt{\{food\_item\}}, and what ingredients do they contain?”
\end{itemize}

\subsection{Mapping Templates to KG Queries}
\label{sec:appendix_mapping}

Each template was programmatically instantiated by substituting entity names from the MMKG. For one-hop templates, we directly queried the KG node properties (e.g., \texttt{caloricValue} for “caloric value” questions). Two-hop templates required traversing relations (e.g., ingredients linked via \texttt{hasIngredient}) and aggregating attributes.

\subsection{Multimodal Response Generation}
\label{sec:appendix_response}

For each generated QA pair, we attached the corresponding image stored in the MMKG:
\begin{itemize}[noitemsep]
  \item Textual answer retrieved via SPARQL queries.
  \item Image retrieved by the \texttt{imagePath} property.
\end{itemize}
This allowed us to produce responses such as:

\begin{quote}
\textbf{Q:} “How do you prepare Masala Dosa?”\\
\textbf{A:} “Combine fermented rice–lentil batter on a hot griddle until golden; fold and serve with chutney.”\\
\textbf{Image:} \includegraphics[height=3cm]{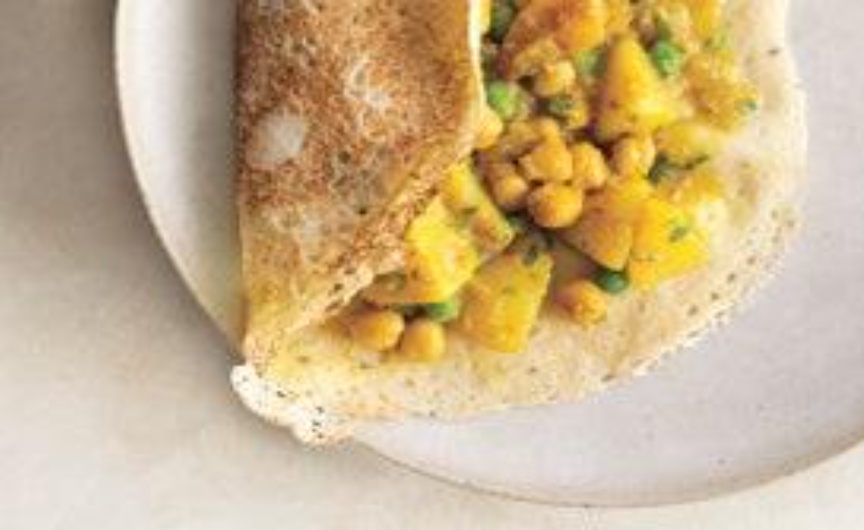}
\end{quote}

\subsection{Automation and System Integration}
\label{sec:appendix_automation}

Our pipeline automates QA generation and response assembly:
\begin{enumerate}[noitemsep]
  \item Instantiate templates with KG entities.
  \item Execute SPARQL queries to fetch answers and image paths.
  \item Post-process outputs: deduplication, clustering-based filtering, and manual validation.
\end{enumerate}
This end-to-end workflow ensures scalability and consistency across the 40K+ QA pairs in the dataset.

\subsection{Qualitative Analysis of QA Samples}
\label{sec:appendix_qualitative}

We manually inspected a random subset of 100 QA pairs to assess linguistic diversity and factual correctness:
\begin{itemize}[noitemsep]
  \item \textbf{Template QA:} Highly precise but repetitive phrasing (e.g., multiple “What is the fat content of…”).
  \item \textbf{LLM-Augmented QA:} Richer variation (e.g., “Can I swap coconut oil for butter in this recipe?”) while maintaining factual alignment.
  \item \textbf{Visual QA:} Questions such as “What color is the sauce?” demonstrated the utility of image-linked queries.
\end{itemize}

These observations confirm that our hybrid approach balances control and diversity, producing QA pairs that are both reliable and varied.


\end{document}